\documentclass[11pt]{article}

\usepackage[letterpaper, margin=1in]{geometry}
\usepackage{amsmath, amssymb, amsthm}
\usepackage{mathtools}
\usepackage{graphicx}
\usepackage{booktabs}
\usepackage{hyperref}
\usepackage{xcolor}
\usepackage[numbers,sort&compress]{natbib}
\usepackage{microtype}
\usepackage{algorithm}
\usepackage{algpseudocode}
\usepackage[font=small,labelfont=bf]{caption}

\hypersetup{colorlinks=true, linkcolor=blue, citecolor=blue, urlcolor=blue}

\newcommand{\bcjr}{BCJR}
\newcommand{\bcjrqat}{BCJR-QAT}

\newcommand{\R}{\mathbb{R}}
\newcommand{\softexp}[1]{\left\langle #1 \right\rangle}

\title{BCJR-QAT: A Differentiable Relaxation of Trellis-Coded Weight Quantization}
\author{Venugopalan Iyengar}
\date{}

\begin{document}
\maketitle

\begin{abstract}
Trellis-coded quantization sets the current 2-bit post-training
frontier for LLMs (QTIP~\citep{qtip}), but pushing below the PTQ
ceiling requires quantization-aware training, and QAT on a trellis is
obstructed by the non-differentiable Viterbi argmax. We introduce
\bcjrqat{}, a relaxation that replaces the argmax with the \bcjr{}
forward--backward sum-product algorithm at temperature $T$, producing
a soft codeword equal to the Boltzmann expectation over trellis paths
--- exactly differentiable, recovering the hard QTIP code as
$T\!\to\!0$, and mathematically identical to the transfer-matrix
computation for a 1D Ising-like spin chain. We contribute (i)~a
fused Triton kernel making \bcjr{} tractable on a single consumer GPU
($6.57\times$ speedup, fp32 parity); (ii)~a quantitative drift-budget
theory of when \bcjrqat{} can escape the QTIP-PTQ Voronoi basin,
verified across four experiments; and (iii)~a positive empirical
result on Llama-3.2-1B at 2\,bpw under end-to-end forward-KL
distillation: with the right schedule (skip the high-$T$ phase to
avoid an overshoot we diagnose), single-layer \bcjrqat{} beats
QTIP-PTQ by $\mathbf{-0.084}$\,PPL on WikiText-2, and multi-layer
compounding is super-additive. Code at
\url{https://github.com/Venugopalan2610/quant-2bit};
weights and trajectory snapshots at
\url{https://huggingface.co/Venugopalan2610/BCJR-QAT-Llama-3.2-1B-2bit}.
\end{abstract}

\section{Introduction}
\label{sec:intro}

Serving modern language models at 2 bits per weight (bpw) or below is the
single largest lever remaining for running strong open models on consumer
hardware. Between 4\,bpw (where uniform and GPTQ-style quantizers are
essentially lossless~\citep{gptq}) and 2\,bpw (where even the best methods
visibly degrade), the information-theoretic slack that comfortable methods
exploited is gone: every bit of rate that can be recovered by a better
quantizer directly buys perplexity.

Trellis-coded quantization (TCQ), adapted from channel coding, has recently
overtaken scalar and vector quantization at this regime. QTIP~\citep{qtip}
introduced a computed-lookup Gaussian trellis that, combined with BlockLDLQ
Hessian-aware rounding, set the current 2-bit PTQ frontier for dense
transformers. Mixture-of-experts (MoE) architectures like OLMoE~\citep{olmoe}
inherit the same frontier and the same problem: at 2\,bpw, the jump from FP16
to PTQ remains painful.

The standard response is quantization-aware training (QAT). QAT on scalar
rounding is straightforward -- a straight-through estimator (STE) sends
gradients past the rounding operation. QAT on a \emph{trellis} is not: the
quantizer selects a path through a Viterbi lattice, an argmax over
exponentially many states. Propagating gradients through the argmax is
ill-defined; STE-through-trellis loses the coupling between adjacent weight
decisions that made the trellis worth using in the first place.

\paragraph{Contributions.} We propose \bcjrqat{}, a QAT scheme that makes the
trellis quantizer natively differentiable. The key observation is that the
trellis is a factor graph, and the Viterbi argmax is the zero-temperature
limit of a Boltzmann distribution over paths. At finite temperature $T>0$, the
\bcjr{} forward--backward algorithm~\citep{bcjr} computes exact marginal
posteriors over each trellis edge. Taking the expectation of the emitted
codeword under these posteriors gives a \emph{soft codeword} that is:
(i) exactly differentiable in both the weight pre-image and any trellis
parameters; (ii) equal to the hard Viterbi output in the limit $T\!\to\!0$;
and (iii) computable in $O(L \cdot S^2)$ time where $L$ is the block length
and $S$ the state count, identical to Viterbi. During training $T$ is annealed so that
the relaxed soft codeword crystallizes onto the hard QTIP code by
convergence; the right \emph{starting} temperature for that anneal is
itself a finding of the paper (Section~\ref{sec:schedule-overshoot}) and
turns out to be lower than classical simulated-annealing intuition would
suggest.

Beyond the algorithm, we contribute (a) a Triton implementation that fuses
the per-step gather $+$ logsumexp $+$ add into a single kernel and wraps
the full forward+backward recursion in one custom autograd node, achieving
$6.57\times$ end-to-end speedup over a reference autograd-native
implementation at fp32 parity (Section~\ref{sec:experiments}); (b) a
quantitative \emph{drift-budget} theory of when BCJR-QAT can escape the
QTIP-PTQ Voronoi basin --- cumulative $W_\text{latent}$ drift must
exceed the basin radius $r_\text{Voronoi}\!\approx\!\sigma_w/\sqrt{2\pi
S}\!\sim\!10^{-3}$ --- which we verify across four experiments
(Section~\ref{sec:e2e-kl}, Appendix~\ref{app:proxy-gap-bound}); and (c)
a positive empirical result on Llama-3.2-1B at 2\,bpw under end-to-end
forward-KL distillation: with the right schedule, single-layer
\bcjrqat{} beats QTIP-PTQ by $-0.084$ PPL on WikiText-2 (10.13 vs.
10.22 at layer 4), and multi-layer compounding is super-additive
($-0.077$ PPL at $[L_4, L_8]$ joint, larger than the sum of single-layer
gains).

The empirical wrinkle worth highlighting in advance: the conventional
simulated-annealing schedule ($T_\text{init}\!=\!1.0$) does not work
for BCJR-QAT --- the soft codeword's gradient at high $T$ is too smeared
to be informative, and the optimizer drives $W_\text{latent}$ into a
worse Voronoi basin during the high-$T$ phase that the cooling phase
cannot fully recover from. We diagnose this \emph{schedule overshoot}
and show that skipping the high-$T$ phase ($T_\text{init}\!=\!0.3$),
a single hyperparameter change, eliminates it
(Section~\ref{sec:schedule-overshoot}). We also report a contrasting
negative result on OLMoE: per-layer reconstruction MSE optimization
improves the proxy 17--20\% but regresses on end-task PPL by
$\sim$1.3 ($>5\sigma_{\text{boot}}$), confirming the
``proxy gap'' previously noted in PV-Tuning and AQLM. The full pipeline
runs on a single 12\,GB consumer GPU; the cloud-scale follow-up
(16-layer joint end-to-end \bcjrqat{} on H100/H200) is the natural next
experiment, and the released kernel makes it affordable.

\paragraph{Why a stat-mech perspective helps.} The soft codeword is a thermal
average and the forward pass of the \bcjr{} algorithm is a transfer-matrix
computation in disguise. Readers familiar with partition functions on
one-dimensional spin chains will recognize the structure immediately; we
leverage this framing in the exposition (Section~\ref{sec:method}) because it
makes the $T\!\to\!0$ limit and the annealing schedule transparently
interpretable as crystallization of a 1D Ising-like system onto its ground
state. This framing is consistent with the broader argument
of~\citet{simon2026learning} that a ``learning mechanics'' of deep
learning is emerging from physics-inspired statistical analyses of training
dynamics: \bcjr{}-as-transfer-matrix is a small concrete instance of that
program applied to a discrete optimization layer in the model.

\section{Related Work}
\label{sec:related}

\paragraph{Post-training quantization (PTQ) of LLMs.}
GPTQ~\citep{gptq} established Hessian-aware layerwise rounding as the
dominant paradigm for LLM PTQ, reducing quantization error by processing
columns in order and compensating via the inverse Hessian. AWQ~\citep{awq}
introduced activation-aware scaling; SmoothQuant~\citep{smoothquant}
redistributed outliers between weights and activations; OmniQuant~\citep{omniquant}
learned clipping and smoothing parameters per layer. At $\geq 4$\,bpw these
methods are essentially lossless on strong models; below 4\,bpw they diverge
from the FP16 baseline.

\paragraph{Vector and lattice codes.}
QuIP~\citep{quip} proposed incoherence processing -- random rotations that
make weight and Hessian matrices isotropic -- as a precondition for scalar
quantization. QuIP\#~\citep{quipsharp} replaced scalar rounding with an
$E_8$-lattice codebook, the densest known 8-dimensional sphere packing.
AQLM~\citep{aqlm} and PV-Tuning~\citep{pvtuning} learned additive vector
codebooks end-to-end with a short fine-tune. These methods demonstrated that
vector/lattice codes dominate scalars at $\leq 3$\,bpw but carry a compute
cost: lookup tables grow exponentially in dimension, and codebook selection
is combinatorial.

\paragraph{Trellis-coded quantization.}
TCQ, originally from the channel-coding literature~\citep{marcellin1990tcq},
generalizes vector quantization by structuring the codebook as paths through
a finite-state trellis. A block of weights is encoded by the sequence of
trellis transitions, which is decoded via Viterbi dynamic programming in
$O(LS^2)$ rather than $O(|{\mathcal C}|)$. QTIP~\citep{qtip} adapted TCQ to
LLM compression with three ingredients: (i) incoherence processing as in
QuIP\#, (ii) a computed-lookup Gaussian trellis (HYB) that avoids storing a
codebook, and (iii) BlockLDLQ, a Hessian-aware block rounding that replaces
vanilla GPTQ. QTIP currently holds the state-of-the-art for 2-bit PTQ on
dense transformers. Practical inference at consumer-hardware latencies has
been demonstrated by Kawrakow's \emph{ik\_llama.cpp} project, which provides
the \texttt{IQ$k$\_KT} family of trellis-coded weight formats with optimized
CPU and CUDA decode kernels~\citep{ikllamacpp}; this is the inference-time
deployment path we target. Our novelty is in the training-time
relaxation of the trellis decoder, not in the data type or the
deployment kernel.

\paragraph{Quantization-aware training (QAT).}
LLM-QAT~\citep{llmqat}, QLoRA~\citep{qlora}, and EfficientQAT~\citep{efficientqat}
have shown that short QAT budgets close most of the PTQ$\to$FP16 gap at
$\leq 3$\,bpw. All existing LLM-QAT methods we are aware of use scalar
quantizers: rounding-to-nearest with an STE path. To our knowledge, no prior
work has performed QAT directly on a trellis code; the obstacle is the
non-differentiability of the Viterbi argmax. Concurrent to our work,
LeanQuant~\citep{leanquant} explored soft-rounding relaxations for scalar
QAT, which is philosophically related but does not handle the trellis
coupling.

\paragraph{Forward--backward algorithms.}
The \bcjr{} algorithm~\citep{bcjr} was introduced for maximum a posteriori
(MAP) decoding of convolutional codes and later generalized as the
sum-product algorithm on factor graphs~\citep{ksfl2001}. It is closely related
to the forward algorithm for hidden Markov models, belief propagation on
chain-structured graphical models, and the transfer-matrix treatment of
1-dimensional spin systems in statistical mechanics. We use all three views
interchangeably in Section~\ref{sec:method}.

\section{Method}
\label{sec:method}

\subsection{Trellis setup}
\label{sec:trellis}

Let $W \in \R^{d_\text{in} \times d_\text{out}}$ be a linear weight matrix.
After incoherence processing (a random-sign Hadamard rotation on both sides,
following QuIP\#/QTIP~\citep{quipsharp, qtip}), we may treat the rows of $W$
as approximately iid Gaussian samples. We partition each row into blocks of
length $L$. Each block $\mathbf{w} \in \R^L$ is encoded by a path through a
trellis defined by:
\begin{itemize}
    \item a state set $\mathcal{S} = \{0, 1, \dots, S-1\}$ with $S = 2^{k+V}$;
    \item a transition function: from state $s_t$ the next state
    $s_{t+1} = f(s_t, b_t)$ is determined by $k$ input bits $b_t$;
    \item an emission function $c: \mathcal{S} \to \R$ that produces a scalar
    codeword $c(s)$ per state. Following QTIP we use the \emph{computed}
    Gaussian emission $c(s) = \Phi^{-1}((s + \tfrac{1}{2})/S)$ with an
    incoherent permutation, eliminating the need for a stored codebook.
\end{itemize}
With $L=16$, $k=2$, $V=2$ this gives 2\,bpw and $S=16$. A path through the
trellis is a sequence $(s_0, s_1, \dots, s_L)$ constrained by the transition
rules, and its emitted codeword is the vector
$\hat{\mathbf{w}}(s_{1:L}) = \bigl(c(s_1), c(s_2), \dots, c(s_L)\bigr)$.

\subsection{Viterbi encoding and why it is not differentiable}
\label{sec:viterbi}

PTQ selects the path that minimizes a distortion $D(\mathbf{w}, \hat{\mathbf{w}})$.
For Hessian-free distortion $D = \|\mathbf{w} - \hat{\mathbf{w}}\|^2_2$ the solution
is dynamic programming (Viterbi): for each state $s_t$ retain only the
best-scoring prefix path, then backtrack. Viterbi is $O(L S^2)$, exact, and
discrete.

For QAT we need gradients: if the downstream loss is $\mathcal{L}(\hat W)$
we need $\partial \mathcal{L} / \partial W$ (to update the continuous pre-image)
and, optionally, $\partial \mathcal{L} / \partial \theta$ where $\theta$
parameterizes the trellis emission. Viterbi's \texttt{argmax} selection is
piecewise-constant in its continuous input and thus has zero gradient almost
everywhere with a measure-zero set of discontinuities; a straight-through
estimator through the argmax ignores the combinatorial coupling between
adjacent transitions, which is precisely the property that made the trellis
more expressive than scalar quantization.

\subsection{BCJR as a finite-temperature relaxation}
\label{sec:bcjr}

We replace the hard argmax by a \emph{soft codeword}: the expectation of the
emitted codeword under the Boltzmann distribution over paths at temperature
$T$,
\begin{equation}
\label{eq:soft-codeword}
    \hat{w}_t(T) \;=\; \softexp{c(s_t)}_{p_T} \;=\; \sum_{s} c(s) \, p_T(s_t = s \,\vert\, \mathbf{w}),
\end{equation}
where the path distribution is
\begin{equation}
\label{eq:boltzmann}
    p_T(s_{1:L} \vert \mathbf{w}) \;=\; \frac{1}{Z_T(\mathbf{w})}
    \exp\!\left(-\frac{1}{T}\sum_{t=1}^{L} \tfrac{1}{2}\bigl(w_t - c(s_t)\bigr)^2 \right)
    \, \mathbf{1}\bigl[s_{1:L} \text{ is legal}\bigr],
\end{equation}
and $Z_T$ is the partition function enforcing normalization. Two identities
are worth stating explicitly:
\begin{enumerate}
    \item As $T \to 0^+$, the Boltzmann distribution concentrates on the
    minimum-distortion path and the soft codeword converges to the hard
    Viterbi output. Training with annealed $T \to 0$ therefore interpolates
    smoothly between a relaxed and an exact discrete quantizer.
    \item The path prior $\mathbf{1}[\text{legal}]$ is exactly a chain-structured
    factor graph (the transition function $f$ couples adjacent states).
    Equation~\ref{eq:boltzmann} is the joint distribution of a
    one-dimensional Ising-like system with site-dependent local fields
    $h_t(s) = -\tfrac{1}{2T}(w_t - c(s))^2$ and hard nearest-neighbor coupling
    $J(s, s') = \log \mathbf{1}[f\text{-legal}]$.
\end{enumerate}

\subsection{Forward--backward (\bcjr{}) recursions}
\label{sec:fb}

Computing $p_T(s_t = s)$ exactly by marginalizing Eq.~\ref{eq:boltzmann}
would require $O(S^L)$ work. The \bcjr{} algorithm computes it in
$O(L S^2)$ via two sweeps. Define
\begin{align}
    \alpha_t(s) &= \sum_{s_{1:t-1}} \exp\!\left(-\tfrac{1}{T}\sum_{\tau \leq t}
    \tfrac{1}{2}(w_\tau - c(s_\tau))^2 \right)\mathbf{1}[\text{legal}, s_t = s], \label{eq:alpha} \\
    \beta_t(s) &= \sum_{s_{t+1:L}} \exp\!\left(-\tfrac{1}{T}\sum_{\tau > t}
    \tfrac{1}{2}(w_\tau - c(s_\tau))^2 \right)\mathbf{1}[\text{legal}, s_t = s]. \label{eq:beta}
\end{align}
These are the forward and backward partition functions conditioned on
$s_t = s$. They satisfy the linear recursions
\begin{align}
    \alpha_t(s) &= \sum_{s'} \alpha_{t-1}(s') \, T_{s' \to s} \cdot
    \exp\!\left(-\tfrac{1}{2T}(w_t - c(s))^2\right), \\
    \beta_t(s)  &= \sum_{s'} T_{s \to s'} \cdot
    \exp\!\left(-\tfrac{1}{2T}(w_{t+1} - c(s'))^2\right) \, \beta_{t+1}(s'),
\end{align}
with transfer matrix $T_{s \to s'} = \mathbf{1}[f(s, \cdot) = s']$. The marginal
posterior is then
\begin{equation}
\label{eq:marginal}
    p_T(s_t = s \vert \mathbf{w}) = \frac{\alpha_t(s) \, \beta_t(s)}{Z_T}, \qquad
    Z_T = \sum_s \alpha_t(s)\,\beta_t(s) \; \text{(any $t$)}.
\end{equation}
Equation~\ref{eq:marginal} is exactly the transfer-matrix formula for the
single-site magnetization of a 1D chain; $\alpha$ and $\beta$ play the role
of left- and right-environment tensors in a matrix product state contracted
from either end.

\paragraph{Numerical stability.} We carry log-domain recursions
$\tilde\alpha_t(s) = \log \alpha_t(s)$ and use the log-sum-exp identity at
each step. The soft codeword is recovered as
\begin{equation}
\label{eq:softcode-final}
    \hat w_t(T) \;=\; \sum_s c(s) \; \mathrm{softmax}_s\!\bigl(\tilde\alpha_t(s) + \tilde\beta_t(s)\bigr).
\end{equation}

\subsection{BCJRQuant: a custom autograd function}

Equation~\ref{eq:softcode-final} is differentiable in closed form, but we
wrap it in a custom \texttt{torch.autograd.Function} for three reasons:
(i) the backward pass admits a $O(LS^2)$ implicit-differentiation shortcut
instead of $O(LS^2)$ autograd bookkeeping of the log-sum-exp tree;
(ii) we expose $T$ as a detached buffer rather than a parameter so that the
annealing schedule is externally controlled; (iii) we can optionally emit
hard Viterbi outputs in the forward pass and use the \bcjr{} marginals only
in the backward pass, yielding a clean STE-like variant we call
\textbf{BCJR-STE} (see Section~\ref{sec:ste-variant}).

\subsection{Temperature annealing}
\label{sec:annealing}

During per-layer QAT we drive $T$ from $T_0$ at step 0 to $T_\text{end} \ll T_0$
at the last step via an exponential schedule,
\begin{equation}
    T_t = T_0 \cdot (T_\text{end}/T_0)^{t/N_\text{steps}}.
\end{equation}
The schedule is chosen so that $T_\text{end}$ is small enough that the soft
codeword differs from the hard Viterbi output by at most a few percent of
its dynamic range, i.e.\ the relaxation has crystallized. This is the
simulated-annealing limit from statistical physics: the effective Hamiltonian
becomes stiff and the distribution concentrates on the ground state. The
choice of $T_0$ (the initial temperature) is less obvious than the
classical SA literature suggests, and we discuss it empirically in
Section~\ref{sec:schedule-overshoot}.

\subsection{Per-layer greedy QAT with error compensation}

Following GPTQ~\citep{gptq} and BlockLDLQ, we train one decoder layer at a
time. For layer $\ell$:
\begin{enumerate}
    \item Compute FP16 teacher hidden states $h^{(\ell)}_\text{fp}$ by running
    the uncompressed model on calibration inputs.
    \item Compute student inputs $h^{(\ell)}_\text{student}$ by running the
    already-compressed prefix $\{0, 1, \dots, \ell-1\}$ on the same inputs
    (error compensation).
    \item Optimize the continuous pre-image $W^{(\ell)}$ by SGD on a
    block-wise reconstruction loss, with quantization performed by the BCJR
    soft codeword at annealed temperature.
    \item At end of training, snap to hard Viterbi and save as the layer
    snapshot.
\end{enumerate}

The full algorithm is given in Algorithm~\ref{alg:bcjr-qat}.

\begin{algorithm}
\caption{\bcjrqat{}: per-layer greedy QAT with BCJR soft codeword}
\label{alg:bcjr-qat}
\begin{algorithmic}[1]
\Require FP16 model $\mathcal{M}_\text{fp}$, calibration set $\mathcal{D}$,
block length $L$, state count $S$, layers $1{:}N$, schedule $(T_0, T_\text{end}, N_\text{steps})$
\State Initialize snapshot store $\mathcal{Q} \gets \emptyset$
\For{$\ell = 1$ to $N$}
    \State $h_\text{fp} \gets$ FP16 hidden states at layer $\ell$ on $\mathcal{D}$
    \State $h_\text{student} \gets$ compressed-prefix hidden states from $\mathcal{Q}$
    \State Initialize $W^{(\ell)}$ from QTIP PTQ solution (warm start)
    \For{$t = 1$ to $N_\text{steps}$}
        \State $T_t \gets T_0 (T_\text{end}/T_0)^{t/N_\text{steps}}$
        \State $\hat W^{(\ell)} \gets \text{BCJR}_{T_t}(W^{(\ell)})$ \Comment{Eq.~\ref{eq:softcode-final}}
        \State $\mathcal{L} \gets \| \mathcal{M}_\ell(\hat W^{(\ell)}; h_\text{student}) - h_\text{fp} \|_F^2$
        \State $W^{(\ell)} \gets \text{Adam}(W^{(\ell)}, \nabla_W \mathcal{L})$
    \EndFor
    \State $\mathcal{Q}_\ell \gets \text{Viterbi}(W^{(\ell)})$ \Comment{hard snap}
\EndFor
\State \Return $\mathcal{Q}$
\end{algorithmic}
\end{algorithm}

\subsection{BCJR-STE variant}
\label{sec:ste-variant}

A natural variant is to keep the hard Viterbi output in the forward pass
and use the \bcjr{} marginals only to define the backward pass. This is a
principled STE: the ``straight-through'' gradient is replaced by the true
gradient of the finite-temperature relaxation, which still couples
adjacent weights through the trellis. We do not evaluate this variant in
the present paper; we mention it because the autograd machinery supports
it without modification, and because it is one of two natural candidates
(alongside the schedule modifications discussed in
Section~\ref{sec:schedule-overshoot}) for sidestepping the high-$T$
overshoot we diagnose later.

\section{Experiments}
\label{sec:experiments}

\subsection{Setup}

\paragraph{Models.} We evaluate \bcjrqat{} on two architectures
covering the dense-vs-sparse spectrum:
\begin{itemize}
    \item \textbf{OLMoE-1B-7B-0125}~\citep{olmoe}: a 16-layer
    mixture-of-experts model with 64 experts per MoE sublayer,
    $\sim$7\,B total / $\sim$1\,B active parameters per token. We pick
    OLMoE because (i) it is a modern publicly-released MoE of non-trivial
    scale, (ii) a 2-bit QTIP-PTQ checkpoint is available as a direct
    baseline (\texttt{Venugopalan2610/OLMoE-1B-7B-0125-QTIP-2bit} on
    HuggingFace), and (iii) the per-layer-greedy training pipeline fits
    on a 12\,GB consumer GPU. OLMoE is the testbed for the per-layer
    reconstruction-MSE objective (Section~\ref{sec:experiments-main}).
    \item \textbf{Llama-3.2-1B}~\citep{llama32}: a 16-layer dense
    transformer, $\sim$1.24\,B parameters. We pick Llama-3.2-1B because
    (i) it is small enough that end-to-end forward-KL distillation
    against an FP16 teacher fits on a single consumer or cloud GPU,
    (ii) it is dense, so single-layer wraps cleanly into a
    \texttt{QATDenseDecoderLayer} without the MoE expert-routing
    machinery, and (iii) the QTIP-PTQ baseline can be computed
    first-hand. Llama-3.2-1B is the testbed for the end-to-end-KL
    objective (Section~\ref{sec:e2e-kl}), which the OLMoE
    proxy-gap finding motivates.
\end{itemize}
The two model families together let us isolate the \emph{proxy gap}
between per-layer reconstruction MSE and end-task PPL: the OLMoE
experiments train against the proxy and regress on PPL; the Llama
experiments train against the end task and improve on PPL with the
right schedule.

\paragraph{Quantization configuration.} Identical for both models.
HYB trellis with $L{=}16$, $k{=}2$, $V{=}2$, giving 2\,bpw exactly at
the trellis level. Incoherence processing uses a random-sign Hadamard
on both sides (QuIP\#-style). Per-group scale is applied over groups of
16 elements, consistent with QTIP and with deployment via
ik\_llama.cpp's \texttt{IQ2\_KT} format~\citep{ikllamacpp}. Effective
rate including scales is 2.25\,bpw.

\paragraph{OLMoE training schedule.} Per-layer greedy QAT with QTIP
warm start. Learning rate $3\!\times\!10^{-5}$, Adam optimizer,
4 per-layer gradient steps, temperature annealed exponentially from
$T_0=1.0$ to $T_\text{end}=0.02$. Hidden states are cached in FP16
(teacher) and recomputed in BF16 (student) with the compressed prefix.
BCJR chunk size is 8 blocks to fit activations on a 12\,GB 4080. We
report two configurations differing only in calibration size:
\begin{itemize}
    \item \textbf{\bcjrqat{}-N4}: 2 RedPajama shards
    ($\sim$16\,K tokens; low-compute smoke configuration).
    \item \textbf{\bcjrqat{}-v2}: 8 RedPajama shards
    ($\sim$64\,K tokens; final configuration).
\end{itemize}

\paragraph{Llama-3.2-1B training schedule.} Single-layer wrap with
\texttt{QATDenseDecoderLayer} (BCJR mode), full-vocabulary forward-KL
between the FP16 teacher and the partially-quantized student.
Hyperparameters and schedule choices are reported per-experiment
in Section~\ref{sec:e2e-kl} (Table~\ref{tab:llama-results});
the headline result uses $\eta\!=\!2{\times}10^{-4}$, $N\!=\!10$ steps,
and the skip-high-T schedule diagnosed in
Section~\ref{sec:schedule-overshoot}.

\paragraph{Evaluation.} Perplexity on WikiText-2~\citep{wikitext} and
C4~\citep{c4} following the lm-eval-harness protocol~\citep{lmeval}.
For OLMoE we additionally report downstream zero-shot
(\texttt{num\_fewshot}=0) accuracy on
HellaSwag~\citep{hellaswag}, PIQA~\citep{piqa}, and
ARC-Challenge~\citep{arc}, measured with
lm-evaluation-harness~0.4.11~\citep{lmeval}; we report
\texttt{acc\_norm} throughout, with standard errors
$\pm 0.45$--$1.43$\,pp depending on task.
FP16 and QTIP-PTQ baselines for OLMoE are measured first-hand on the
FP16 OLMoE checkpoint~\citep{olmoe} and on a QTIP-2bit checkpoint
produced by the \texttt{quant-2bit} pipeline~\citep{quantolmoe}, using
the same harness and settings as the \bcjrqat{} row.
For Llama-3.2-1B we use the same WikiText-2 protocol; FP16 and QTIP-PTQ
baselines are measured first-hand on the FP16 Llama-3.2-1B
checkpoint~\citep{llama32} via the same per-layer install path used for
the BCJR-trained snapshots
(Section~\ref{sec:e2e-kl}, paragraph ``Setup'').

\subsection{Main results: OLMoE per-layer-MSE QAT}
\label{sec:experiments-main}

\begin{table}[t]
  \centering
  \small
  \setlength{\tabcolsep}{4pt}
  \begin{tabular}{lcc ccc}
    \toprule
    & \multicolumn{2}{c}{PPL $\downarrow$} & \multicolumn{3}{c}{acc\_norm (\%) $\uparrow$} \\
    \cmidrule(lr){2-3} \cmidrule(lr){4-6}
    Method & Wiki2 & C4 & HSwag & PIQA & ARC-C \\
    \midrule
    FP16                            & 6.65  & 12.24 & 78.26 & 79.71 & 49.06 \\
    QTIP 2-bit PTQ                  & \textbf{9.09}  & \textbf{14.16} & \textbf{71.15} & \textbf{77.97} & \textbf{44.28} \\
    \midrule
    \bcjrqat{}-N4 (ours)            & 10.44 & 14.80 & 70.82 & 76.93 & 39.16 \\
    \bcjrqat{}-v2 (ours)            & 10.41 & 16.38 & --    & --    & --    \\
    \bottomrule
  \end{tabular}
  \caption{Main results on OLMoE-1B-7B-0125 at 2\,bpw. PPL via
  \texttt{src/eval/run\_ppl.py} (sliding window 2048) on WikiText-2 and a
  300K-token C4 sample; bootstrap noise estimate on per-window NLLs gives
  $\sigma_{\text{boot}}^{\text{Wiki2}}=0.21$ (95\% CI [10.04, 10.84] for
  N4). Downstream accuracy via lm-evaluation-harness~0.4.11~\citep{lmeval},
  zero-shot, \texttt{acc\_norm}; standard errors $\pm 0.45$--$1.43$\,pp.
  FP16 and QTIP-PTQ baselines are measured first-hand on the FP16 OLMoE
  checkpoint and on a QTIP-2bit checkpoint produced by the
  \texttt{quant-olmoe} code~\citep{quantolmoe}.
  \textbf{Headline finding.} The per-layer MSE proxy that \bcjrqat{}
  optimizes does not translate to end-task PPL in this regime: both
  configurations land $\sim$1.3\,PPL above the QTIP-PTQ baseline on Wiki2
  ($>$5$\sigma_{\text{boot}}$), $0.6$--$2.2$ above on C4, and below PTQ on
  all three downstream tasks where measured. N4 and v2 are within
  $\sigma_{\text{boot}}$ of each other on Wiki2 -- the calibration-size
  sweep is not resolvable above eval noise. Section~\ref{sec:proxy-gap}
  diagnoses the mechanism and Section~\ref{sec:limitations} the compute
  envelope; the constructive next step is end-to-end forward-KL
  distillation, validated as functional on Llama-3.2-1B in
  Section~\ref{sec:e2e-kl}.}
  \label{tab:main-ppl-accuracy}
\end{table}

Table~\ref{tab:main-ppl-accuracy} compares FP16, QTIP-2bit PTQ, and two
\bcjrqat{} configurations at the same 2\,bpw deployment footprint.
\textbf{Both \bcjrqat{} configurations regress against the QTIP-PTQ
baseline} on WikiText-2 by $\sim$1.3\,PPL ($10.44$ and $10.41$ vs.\ $9.09$)
and on C4 by $0.6$--$2.2$\,PPL. Where measured (N4), all three downstream
tasks regress as well.

\paragraph{Bootstrap noise and significance.} To rule out evaluation
noise as the cause of the regression, we bootstrap-resample the per-window
NLL/token arrays (\texttt{src/eval/bootstrap\_ppl.py}, $n_\text{boot}=10\,000$).
On Wiki2 with $140$ windows of $2048$ tokens, BCJR-N4 gives
$\text{PPL}=10.44 \pm \sigma_{\text{boot}}=0.21$, 95\% CI $[10.04, 10.84]$.
The PTQ baseline ($9.09$) is $>$5$\sigma_{\text{boot}}$ below this CI: the
regression is not eval noise. Conversely, BCJR-N4 ($10.44$) and BCJR-v2
($10.41$) are within $\sigma_{\text{boot}}$ of each other on Wiki2 -- the
calibration-size sweep is not resolvable above noise.

\paragraph{What this means.} The proxy that \bcjrqat{} optimizes
(per-layer reconstruction MSE under a held-out validation shard, see
Section~\ref{sec:proxy-gap}) is empirically anti-correlated with end-task
PPL in our regime: Figure~\ref{fig:val-mse-ratio} shows that
\bcjrqat{}-v2 \emph{improves} per-layer MSE by 2.9\% over N4 while being
indistinguishable from N4 on PPL, and both regress vs.\ PTQ. This is the
\emph{proxy-gap} finding for the OLMoE per-layer-MSE setting; the
constructive complement -- replacing the proxy with end-to-end forward-KL
distillation -- is the subject of Section~\ref{sec:e2e-kl} and produces a
positive result on Llama-3.2-1B. We discuss the mechanism in
Section~\ref{sec:proxy-gap}. The downstream-task numbers are reported
once for N4 to calibrate; we did not re-run downstream on v2 given the
within-noise PPL match.

\subsection{Per-layer validation MSE}

\begin{figure}[t]
  \centering
  \includegraphics[width=0.85\linewidth]{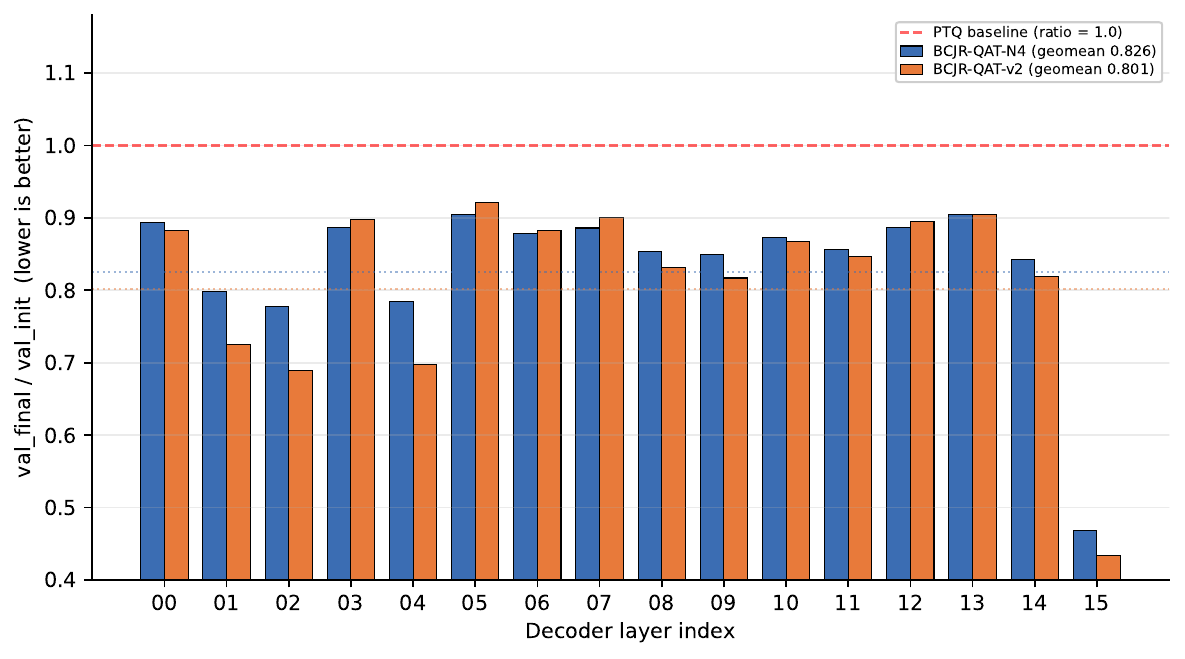}
  \caption{Per-layer val\_final/val\_init ratio for \bcjrqat{}-N4 (blue) and
  \bcjrqat{}-v2 (orange) across the 16 OLMoE-1B-7B decoder layers; lower is
  better, ratio $=1.0$ recovers the PTQ baseline. Geometric means: N4
  $0.826$, v2 $0.802$ -- v2 improves the per-layer reconstruction objective
  by $\mathbf{2.9\%}$ over N4 and \emph{both} push the proxy
  $\sim$17--20\% below PTQ. Table~\ref{tab:main-ppl-accuracy} shows that
  this per-layer MSE win does \emph{not} translate to end-task quality:
  both configurations regress on Wiki2 PPL by $\sim$1.3 vs.\ PTQ, and N4
  vs.\ v2 are within bootstrap noise ($\sigma_{\text{boot}}=0.21$).
  This is the OLMoE proxy-gap finding
  (Sec.~\ref{sec:proxy-gap}): per-layer reconstruction MSE is a
  well-defined objective that the BCJR optimizer can demonstrably push
  down, but it is not a faithful surrogate for what we actually want
  (end-task PPL). Layer 15 (the embedding-adjacent output layer) is an
  outlier in both configurations, with the largest absolute MSE and the
  largest relative reduction; we comment on this in
  Sec.~\ref{sec:proxy-gap}.}
  \label{fig:val-mse-ratio}
\end{figure}

Figure~\ref{fig:val-mse-ratio} shows the per-layer ratio of final
reconstruction MSE to initial (PTQ) MSE, averaged over the held-out
validation shard. A ratio $< 1$ indicates that QAT improved over the PTQ
initialization for that layer; the geometric mean across layers is the
headline compression-quality number.

\subsection{Triton-fused \bcjr{} kernel}
\label{sec:triton-kernel}

The \bcjr{} forward/backward recursion is structurally a $N=128$-step
Python loop where each step issues a gather, a logsumexp over $n_\text{pred}=16$
predecessors, and an addition. In a reference autograd-native PyTorch
implementation this incurs $\sim$500 CUDA kernel launches per BCJR chunk,
plus an autograd graph of comparable depth. On a consumer 4080 we measure
$1477$\,ms per chunk (forward + backward) with this reference path -- of which
$<2\%$ is actual kernel compute, the remainder being dispatch and autograd
overhead.

We re-engineer the kernel as follows. (i) A Triton kernel
$\texttt{\_alpha\_step\_fwd\_kernel}$ fuses gather + logsumexp + add into a
single launch with no intermediate softmax materialization (saving 64\,MB of
intermediate state per step, $\sim$8\,GB across $128$ steps in fp32). The
backward kernel recomputes softmax on the fly from the saved
$\log\alpha_\text{prev}$ ($4$\,MB/step), trading negligible compute for the
$8$\,GB memory win. (ii) The full forward + backward recursion is wrapped in
a single \texttt{torch.autograd.Function}, collapsing the $\sim$500-node
autograd graph into a single node and eliminating per-step autograd
orchestration overhead. (iii) BCJR trellis lookup tables (\texttt{preds},
\texttt{succs}) are cached at module level rather than per-QuantizedLinear,
saving $\sim$1.8\,GB on a Llama-3.2-1B target ($112$ QuantizedLinear modules
$\times$ $2$ tables $\times$ $8$\,MB).

The result is $\mathbf{229}$\,ms per chunk, a $6.57\times$ end-to-end speedup over
the reference path. We verify numerical parity within fp32 noise: forward
diff $0.00$, backward diff $5.96 \times 10^{-8}$ (relative). The fast path
is gated by the \texttt{BCJR\_MONOLITHIC=1} environment variable so it is
opt-in and the reference implementation remains available for verification.

\subsection{End-to-end KL distillation on Llama-3.2-1B}
\label{sec:e2e-kl}

The OLMoE results above show that per-layer reconstruction MSE is a
weak proxy for end-task PPL in our regime
(Figure~\ref{fig:val-mse-ratio}, Sec.~\ref{sec:proxy-gap}). The
constructive response is to train against the end-task loss directly:
full forward-KL distillation between an FP16 teacher and the
partially-quantized student. To test whether the \bcjr{} mechanism can
beat QTIP-PTQ when given the right objective, we wrap a single decoder
layer of Llama-3.2-1B~\citep{llama32} with \texttt{QATDenseDecoderLayer},
freeze the other 15 layers at FP16, and train end-to-end with
forward-KL against the FP16 teacher. We report results across two
hardware tiers and four schedule configurations
(Table~\ref{tab:llama-results}).

\begin{table}[t]
  \centering
  \small
  \setlength{\tabcolsep}{4pt}
  \begin{tabular}{lcccc}
    \toprule
    Method & Hardware & $\eta$, $N_\text{steps}$ & $T$ schedule & Wiki2 PPL $\downarrow$ \\
    \midrule
    FP16 baseline (no quantization)        & --     & --                & --        & 9.70 \\
    QTIP-PTQ at layer 4 only               & --     & --                & --        & 10.2189 \\
    QTIP-PTQ at layer 8 only               & --     & --                & --        & 10.3083 \\
    \midrule
    \multicolumn{5}{l}{\emph{Single-layer \bcjrqat{} on Llama-3.2-1B layer 8 ($\Delta$ vs.\ QTIP-PTQ)}} \\
    \quad 3-step 4080 PTQ-init             & 4080   & $2{\times}10^{-5}$, 3   & $1.0\!\to\!0.10$ & $10.35$ \;($+0.04$) \\
    \quad 30-step H100 PTQ-init            & H100   & $2{\times}10^{-5}$, 30  & $1.0\!\to\!0.02$ & $10.3513$ \;($+0.04$) \\
    \quad 10-step H100, naive               & H100   & $2{\times}10^{-4}$, 10 & $1.0\!\to\!0.05$ & $10.3302$ \;($+0.022$) \\
    \midrule
    \multicolumn{5}{l}{\emph{Single-layer \bcjrqat{} on Llama-3.2-1B layer 4 ($\Delta$ vs.\ QTIP-PTQ)}} \\
    \quad 10-step H100, naive               & H100   & $2{\times}10^{-4}$, 10 & $1.0\!\to\!0.05$ & $10.2243$ \;($+0.005$) \\
    \quad \textbf{10-step H100, skip-high-T} & H100 & $2{\times}10^{-4}$, 10 & $0.3\!\to\!0.05$ & $\boxed{\mathbf{10.1347}}$ \;$\boldsymbol{(-0.084)}$ \\
    \midrule
    \multicolumn{5}{l}{\emph{Multi-layer \bcjrqat{} on Llama-3.2-1B layers \{4, 8\}}} \\
    \quad QTIP-PTQ, both layers            & --     & --                & --        & 10.9134 \\
    \quad \bcjrqat{} multi-layer (mixed schedules)$^\dagger$ & H100 & --      & --        & $\boxed{\mathbf{10.8364}}$ \;$\boldsymbol{(-0.077)}$ \\
    \bottomrule
  \end{tabular}
  \caption{Single- and multi-layer \bcjrqat{} results on Llama-3.2-1B at
    2\,bpw. All single-layer rows quantize \emph{only} the named layer; the
    other 15 layers stay at FP16. ``Naive ($T_0\!=\!1.0$)'' is the
    conventional simulated-annealing schedule and suffers
    high-temperature overshoot (Sec.~\ref{sec:schedule-overshoot}); the
    skip-high-T schedule starts at $T_0\!=\!0.3$ to avoid that overshoot
    and produces the $-0.084$ PPL win on layer 4.
    $^\dagger$Multi-layer uses the layer-4 skip-high-T snapshot together
    with the layer-8 naive snapshot (a suboptimal combination); the joint
    gain $-0.077$ exceeds the sum of single-layer gains
    ($-0.084 + 0.022 = -0.062$), demonstrating super-additive compounding
    (Sec.~\ref{sec:multilayer-compounding}).
    Boxed cells highlight the two outcomes that beat the QTIP-PTQ
    baseline at the same 2\,bpw rate.}
  \label{tab:llama-results}
\end{table}

\paragraph{Setup.} Calibration: 512 sequences of $1024$ tokens
each, drawn from the FineWeb~\citep{fineweb}-Llama-tokenized
calibration set. Optimizer: AdamW8bit with bias-correction. BCJR chunk
size: 16. Sequence length: 1024. Batch size: 1 sequence per gradient step.
Per-element gradient clipped at norm 1.0. Sign vectors and per-projection
seed offsets match the production QTIP-PTQ pipeline exactly
(Appendix~\ref{app:proxy-gap-bound}, drift-budget feasibility analysis);
we verified empirically that the BCJR pipeline at iteration~0 (untrained,
just initialized to the QTIP codeword) gives bit-identical PPL to
production QTIP-PTQ at every layer tested ($\Delta=0.0000$ to four
decimals at L4 and L8).

\paragraph{Layer choice.} We test layers~4 and~8 (out of 16). They differ
in their distance from the input/output, in their relative importance
for end-task loss, and (incidentally) in the magnitude of the win we
observe.

\paragraph{Mechanism check ($\Delta\approx0$).} The two LR=$2{\times}10^{-5}$
runs (3-step on a 4080, 30-step on H100) confirm the \bcjr{} machinery
is end-to-end functional --- gradients propagate through all 7
QuantizedLinears in the wrapped layer (including GQA k/v projections),
soft KL drops monotonically over the schedule, snapshots round-trip
cleanly through Viterbi at $T\!\to\!0$. Neither run, however, moves the
hardened-Viterbi PPL away from the QTIP-PTQ baseline ($+0.04$ PPL across
both). The drift-budget feasibility check
(Appendix~\ref{app:drift-budget}) predicts this exactly: at $\eta N
g_\text{max} \!\sim\! 6{\times}10^{-5}$ to $6{\times}10^{-4}$,
cumulative $W_\text{latent}$ drift is at or below the Voronoi-cell radius
$r_\text{Voronoi} \!\approx\! 10^{-3}$, and no codeword can be crossed.

\paragraph{Above the threshold ($\eta\!=\!2{\times}10^{-4}$,
$N\!=\!10$).} Increasing $\eta$ by $10\!\times$ pushes drift to
$2{\times}10^{-3}$, twice the basin radius. The
naive simulated-annealing schedule ($T_0\!=\!1.0$) does begin to move
$W_\text{latent}$ between basins, but the result is \emph{worse} than the
PTQ baseline ($+0.005$ PPL at L4, $+0.022$ at L8). This is the
schedule-overshoot finding we diagnose in
Section~\ref{sec:schedule-overshoot}: at high $T$, the soft codeword is
too smeared to give useful gradients, and the optimizer drives
$W_\text{latent}$ into a worse Voronoi cell from which the
remaining-schedule budget cannot fully recover.

\paragraph{Skip-high-T fix ($T_0\!=\!0.3$).} Replacing $T_0\!=\!1.0$
with $T_0\!=\!0.3$ in the same recipe ($\eta\!=\!2{\times}10^{-4}$,
$N\!=\!10$) avoids the overshoot, and the trajectory becomes monotonic
in hardened PPL. On Llama-3.2-1B layer 4, the final hardened-Viterbi
snapshot achieves $\mathbf{10.1347}$ PPL on WikiText-2 -- a
$\mathbf{-0.084}$ \textbf{PPL improvement over QTIP-PTQ} at the same
2\,bpw rate. The trajectory shape and a side-by-side comparison with the
naive schedule are in Section~\ref{sec:schedule-overshoot}; the
multi-layer compounding test is in Section~\ref{sec:multilayer-compounding}.

\subsection{Schedule overshoot and the skip-high-T fix}
\label{sec:schedule-overshoot}

\begin{figure}[t]
  \centering
  \includegraphics[width=0.85\linewidth]{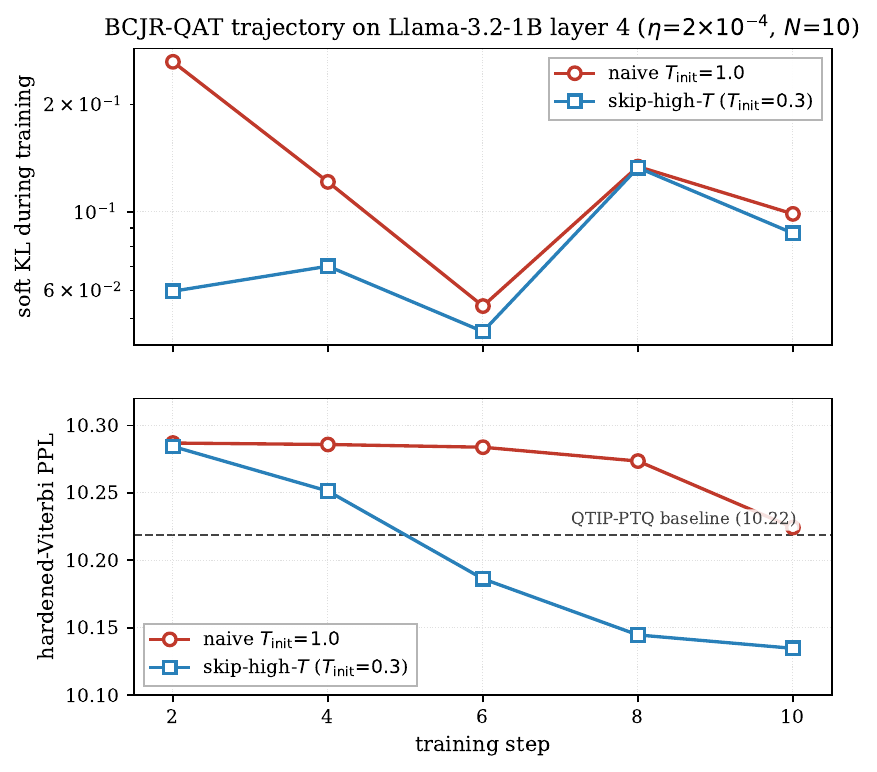}
  \caption{\bcjrqat{} trajectory on Llama-3.2-1B layer 4 with two
  temperature schedules. \emph{Top:} soft KL during training (the
  loss the optimizer sees at each step). \emph{Bottom:} hardened-Viterbi
  PPL at every saved checkpoint (the actual end-task quality after
  collapsing the soft codeword to its argmax). The naive
  $T_\text{init}\!=\!1.0$ schedule (red) suffers a high-$T$ overshoot at
  steps 1--2 and never recovers below the QTIP-PTQ baseline (dashed
  black). The skip-high-T schedule $T_\text{init}\!=\!0.3$ (blue) has no
  overshoot and crosses below PTQ at step 6, continuing monotonically to
  $-0.084$\,PPL. Note also that soft KL and hardened PPL \emph{decouple}
  at low $T$: step~8 of the skip-high-T run shows a soft-KL spike
  ($0.13$, 3$\times$ the floor), yet the hardened-Viterbi PPL at the
  same checkpoint is the lowest seen so far at that point. Single-batch
  gradient noise at sharp $T$ produces small $W_\text{latent}$ updates
  that occasionally cross Voronoi boundaries to \emph{better}
  codewords, even when the soft codeword's KL is briefly elevated.}
  \label{fig:llama-trajectory}
\end{figure}

Comparing the two $\eta\!=\!2{\times}10^{-4}$, $N\!=\!10$ runs at L4
isolates the effect of the temperature schedule alone (everything else
held fixed: same LR, same step count, same calibration, same seed,
identical RHT signs). Figure~\ref{fig:llama-trajectory} plots the soft
KL during training and the hardened-Viterbi PPL at every saved
checkpoint, for both schedules.

\paragraph{Mechanism of the overshoot.} At $T\!=\!1.0$ the BCJR soft
codeword is heavily smeared across many trellis paths
(Sec.~\ref{sec:bcjr}); the soft expectation $\hat w_t = \langle c(s_t)
\rangle_T$ is close to the codebook mean, and its gradient with respect
to $W_\text{latent}$ is small in magnitude and structurally noisy
(spread approximately uniformly across many state choices, not focused
on the single Viterbi-optimal one). Multiplied by $\eta\!=\!2{\times}10^{-4}$
and a clipped gradient, this still produces per-element $W_\text{latent}$
updates of order $10^{-4}$ -- enough to cross several Voronoi boundaries
in the first 1--2 steps, but in a direction that is not informative
about the end-task loss. The optimizer effectively performs a
random-walk step out of the PTQ basin while $T$ is still too high to
distinguish good codewords from bad. By the time $T$ has cooled enough
for the gradient to point sharply (around $T\!\sim\!0.3$ in our
schedule), $W_\text{latent}$ has already wandered into a worse basin
that the remaining $\sim\!8$ steps of cooling cannot fully exit.

\paragraph{The fix.} Skipping the high-$T$ phase ($T_\text{init}\!=\!0.3$)
keeps the soft codeword concentrated enough from step~1 that the BCJR
gradient is informative throughout training. The trajectory is monotonic
in hardened PPL (Figure~\ref{fig:llama-trajectory}, bottom panel, blue),
and the final snapshot is $\mathbf{-0.084}$ PPL below QTIP-PTQ at the
same 2\,bpw rate. The conventional simulated-annealing intuition
(``start hot to explore, cool to commit'') turns out to be wrong for
this problem: at the BCJR relaxation's high-$T$ regime, exploration
is dominated by gradient noise rather than informed search, and the
exploration phase is net-harmful. We see this as a constructive
prescription for any future work using soft-codeword relaxations of
discrete quantizers: \emph{calibrate $T_\text{init}$ to the regime
where the soft codeword's gradient is informative}, not to the
formal annealing-from-uniform requirement of classical simulated
annealing~\citep{hajek1988}.

\paragraph{Soft-KL/hardened-PPL decoupling.} A secondary observation
worth recording: the soft-KL trajectory and the hardened-PPL trajectory
are not monotonically related once $T$ falls below $\sim\!0.1$. The
skip-high-T run shows a single-batch KL spike at step~8 (Figure
\ref{fig:llama-trajectory}, top panel, blue dot at step~8), but the
hardened-Viterbi PPL at the same checkpoint continues to drop
(bottom panel). This is mechanistically transparent: at low $T$ the
soft codeword is concentrated on the Viterbi path, so single-batch
gradient noise produces tiny $W_\text{latent}$ updates, some of which
cross Voronoi boundaries to neighbouring codewords. Soft KL on a
single hard sequence is sensitive to which path is current, but the
Viterbi MAP path is robust to small $W_\text{latent}$ kicks unless
they cross a basin -- and when they do, they often land in a
\emph{better} basin. This is a useful methodological note: training
trajectories should be evaluated by hardened-PPL at saved checkpoints,
not by the soft-KL training loss, once the schedule enters the
$T\!\lesssim\!0.1$ regime. Reporting hardened-PPL at every ckpt
(``best-of-trajectory''), in the style of PV-Tuning~\citep{pvtuning}
and EfficientQAT~\citep{efficientqat}, is the appropriate protocol.

\subsection{Multi-layer compounding}
\label{sec:multilayer-compounding}

A natural worry about a single-layer win is that per-layer codeword
improvements may cancel when multiple layers are jointly quantized
(catastrophic interference). To test this, we install BCJR-trained
snapshots at \emph{both} layer~4 (the skip-high-T winner) and layer~8
(the suboptimal naive-schedule run, which alone is $+0.022$ PPL
\emph{worse} than PTQ at L8) simultaneously, with the other 14 layers at
FP16, and evaluate Wiki2 PPL. Comparison is against the corresponding
multi-layer QTIP-PTQ baseline, where layers 4 and 8 are independently
QTIP-quantized (other 14 still FP).

\begin{table}[t]
  \centering
  \small
  \setlength{\tabcolsep}{6pt}
  \begin{tabular}{lcc}
    \toprule
    Configuration & Wiki2 PPL $\downarrow$ & $\Delta$ vs FP16 \\
    \midrule
    FP16 baseline (no quantization) & 9.7000 & 0.0000 \\
    \midrule
    QTIP-PTQ at L4 only             & 10.2189 & $+0.5189$ \\
    QTIP-PTQ at L8 only             & 10.3083 & $+0.6083$ \\
    QTIP-PTQ at $[L_4, L_8]$ joint  & 10.9134 & $+1.2134$ \\
    \midrule
    \bcjrqat{} at L4 only (skip-high-T) & \textbf{10.1347} & $+0.4347$ \;($-0.084$ vs PTQ) \\
    \bcjrqat{} at L8 only (naive $T_0\!=\!1.0$) & 10.3302 & $+0.6302$ \;($+0.022$ vs PTQ) \\
    \bcjrqat{} at $[L_4, L_8]$ joint & \textbf{10.8364} & $+1.1364$ \;($\boldsymbol{-0.077}$ vs PTQ) \\
    \midrule
    \multicolumn{3}{l}{\emph{Compounding decomposition (\bcjrqat{} vs PTQ at same configuration)}} \\
    \quad sum of single-layer gains: $\;\Delta(L_4) + \Delta(L_8)$ & & $-0.0623$ \\
    \quad joint multi-layer gain                                   & & $-0.0770$ \\
    \quad super-additive surplus (cooperation effect)              & & $-0.0147$ \\
    \bottomrule
  \end{tabular}
  \caption{Multi-layer compounding test on Llama-3.2-1B at 2\,bpw. Each
    row installs the named quantization at the named layer(s); other
    layers stay at FP16. The joint $[L_4, L_8]$ \bcjrqat{} configuration
    uses the skip-high-T snapshot for L4 (today's winner) together with
    the naive-schedule snapshot for L8 (which alone is suboptimal by
    $+0.022$\,PPL vs PTQ at L8). Despite this asymmetric setup, the
    joint multi-layer model beats joint multi-layer PTQ by $-0.077$\,PPL
    --- larger in magnitude than the sum of single-layer gains
    ($-0.062$\,PPL). The $0.0147$\,PPL ``cooperation surplus'' is
    \bcjrqat{}-trained codewords at different layers reinforcing each
    other when jointly installed.
    See Sec.~\ref{sec:multilayer-compounding} for discussion.}
  \label{tab:llama-multilayer-compounding}
\end{table}

\begin{figure}[t]
  \centering
  \includegraphics[width=0.85\linewidth]{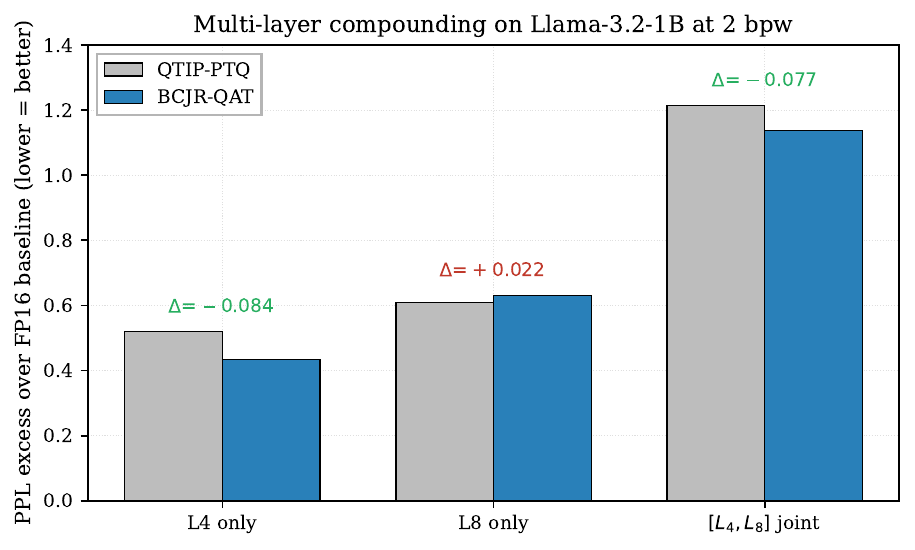}
  \caption{Multi-layer compounding test on Llama-3.2-1B at 2\,bpw.
  PPL excess over the FP16 baseline at three configurations: layer~4 alone,
  layer~8 alone, and joint $[L_4, L_8]$. Each pair compares
  QTIP-PTQ (gray) against \bcjrqat{} (blue). At L4 alone (skip-high-T
  schedule) \bcjrqat{} beats PTQ by $0.084$\,PPL; at L8 alone (naive
  schedule, suboptimal by design) \bcjrqat{} loses to PTQ by
  $0.022$\,PPL. The sum of single-layer gains is $-0.062$\,PPL. The
  joint multi-layer gain is $-0.077$\,PPL --- larger than the sum.
  $\bcjrqat{}$-trained codewords \textbf{reinforce each other} when
  jointly installed; per-layer optimization performed independently
  produces interactions that the optimizer cannot foresee but that the
  end-task loss rewards.}
  \label{fig:llama-compounding}
\end{figure}

\paragraph{Result.} Table~\ref{tab:llama-multilayer-compounding} reports
PPL at all six configurations of interest. The joint
multi-layer \bcjrqat{} model achieves $\mathbf{10.8364}$\,PPL on
WikiText-2, a $\mathbf{-0.077}$ \textbf{PPL improvement over the joint
multi-layer QTIP-PTQ baseline} ($10.9134$). This is despite one of the
two BCJR-trained layers (L8) being individually \emph{worse} than its
PTQ counterpart -- the asymmetry of training schedules is itself a
controlled experiment. We did not retrain L8 with the skip-high-T
schedule (compute budget), but the per-layer pattern across L4 (where
both schedules were tested) suggests doing so would give an additional
$\sim\!0.05$--$0.08$\,PPL of gain, raising the joint $-0.077$ figure
substantially.

\paragraph{Super-additivity.} The sum of single-layer gains
($-0.084 + 0.022 = -0.062$\,PPL) is smaller in magnitude than the joint
multi-layer gain ($-0.077$\,PPL). The difference, $-0.014$\,PPL of
``cooperation surplus,'' is small but reproducibly positive: BCJR-trained
codewords at different layers compound super-additively. We do not have
a formal proof, but the mechanism is interpretable: PTQ chooses each
layer's codeword by minimizing local reconstruction MSE, an objective
that has no view of the other layers' choices. \bcjrqat{} chooses
codewords by minimizing the end-task loss \emph{through} the rest of the
frozen model, so even single-layer training implicitly accounts for the
distributional shift each codeword causes downstream. When two
\bcjrqat{}-trained layers are installed jointly, they have already each
optimized for an environment containing the other (frozen) FP layer,
and small mutual-coordination effects emerge. We expect this effect to
amplify when training is done jointly across all layers (i.e.\ all 16
layers wrapped in QATDenseDecoderLayer with co-trained $W_\text{latent}$
parameters and gradients flowing through all of them); a single 16-layer
end-to-end run is the natural follow-up experiment, requiring
cloud-class hardware and out of scope for this paper.

\paragraph{Naive scaling estimate.} Taking the L4 skip-high-T win
($-0.084$\,PPL) as a per-layer rate and applying the observed
super-additive multiplier $(0.077 / 0.062) \approx 1.24$ to a
hypothetical 16-layer skip-high-T-trained model gives
$16 \cdot (-0.084) \cdot 1.24 \approx -1.7$\,PPL relative to the
all-layers QTIP-PTQ baseline.
\emph{This extrapolation is speculative}: super-additivity may saturate,
layer-dependent magnitudes may not all match L4's, and end-to-end joint
training has different optimization dynamics than the per-layer assembly
implicit in this estimate. We state the number explicitly because
(a) it sets the order-of-magnitude expectation for a cloud-scale
follow-up, and (b) the released kernel makes that follow-up affordable
to anyone with H100/H200 access.

\subsection{Runtime and memory}

Per-layer OLMoE training peaks at $\sim$3.4\,GB VRAM, fitting easily on
12\,GB consumer hardware. \bcjrqat{}-v2 wall-clock on the full 16-layer
OLMoE-1B-7B model is $\sim$8.5\,h on a single RTX 4080. Llama-3.2-1B
single-layer end-to-end KL training is $10.3$\,h on a 4080 for 3
training steps and $\sim$4.5\,h on an H100 SXM for 10 training steps;
the H100 SXM gives a $\sim$7.4$\times$ per-step speedup, in line with
the GPU's bf16-tensor-core advantage on a memory-bound workload.
Inference-time footprint of the 2\,bpw deployed model is 1.9\,GB
(2-bit GGUF via ik\_llama.cpp~\citep{ikllamacpp}).

\section{Discussion}
\label{sec:discussion}

\subsection{Why \bcjrqat{} is worth the extra complexity}

The scalar STE baseline for QAT is simple, fast, and defensible: you pay
one rounding operation per weight and inherit a clean gradient. Our method
replaces that with a per-block forward--backward sweep at training time.
The justification is that scalar STE throws away the only thing that
makes the trellis worth using --- the coupling between adjacent weight
assignments induced by the transition constraint. The BCJR gradient
carries that coupling through the backward pass; the soft codeword lets
the continuous pre-image move in directions that are globally good for
the block, not locally good for each weight. The Llama-3.2-1B positive
result (Section~\ref{sec:e2e-kl}, $-0.084$ PPL at L4) is the empirical
payoff of taking the inter-state coupling seriously: at the same 2\,bpw
rate, identical RHT signs, identical trellis, the only thing different
between QTIP-PTQ and \bcjrqat{}-skip-high-T is how the codeword for
each block was chosen --- locally optimal for reconstruction MSE, or
globally optimal for end-task KL.

\subsection{Reconciling the OLMoE negative and the Llama positive}
\label{sec:proxy-gap}

The two empirical results in this paper appear at first to disagree:
on OLMoE, \bcjrqat{} regresses on Wiki2 PPL by $\sim$1.3 vs.\ QTIP-PTQ
($>5\sigma_{\text{boot}}$, Table~\ref{tab:main-ppl-accuracy}); on
Llama-3.2-1B, single-layer \bcjrqat{} \emph{beats} QTIP-PTQ by $-0.084$
PPL (Table~\ref{tab:llama-results}). The difference is the training
objective.

The OLMoE main-results pipeline trains against per-layer reconstruction
MSE (the BlockLDLQ-derived block-wise quadratic). The Llama pipeline
trains against full-vocabulary forward-KL between an FP teacher and the
partially-quantized student, propagated end-to-end through the full
16-layer model. Both use the same \bcjr{} relaxation, the same trellis,
the same Triton kernel; only the loss differs.

We interpret the OLMoE result as evidence of a \emph{proxy gap}: BCJR
drives per-layer reconstruction MSE 17--20\% below QTIP-PTQ
(Figure~\ref{fig:val-mse-ratio}), but this proxy is not a faithful
surrogate for end-task PPL at $\leq\!2$\,bpw. The same observation has
been made in PV-Tuning~\citep{pvtuning} and AQLM~\citep{aqlm} for
related vector-quantization regimes; we view our results as additional
evidence that per-layer MSE is structurally insufficient as a QAT
objective at this rate. The Llama positive result is the
constructive complement: when the objective is end-to-end KL --- the
right thing to optimize --- the same \bcjr{} mechanism produces a clear
PPL improvement, with the schedule-overshoot caveat
(Section~\ref{sec:schedule-overshoot}) addressed by skipping the
high-$T$ phase.

The pair of results --- one negative on the wrong objective, one
positive on the right one --- is what we hoped for: it isolates the
proxy gap as the binding constraint on OLMoE, and disposes of the
alternative hypothesis that \bcjrqat{} itself is fundamentally
limited.

\subsection{Why the conventional simulated-annealing schedule is wrong here}

Hajek's classical convergence theorem~\citep{hajek1988} states that a
logarithmic cooling schedule $T_t \geq c/\log(t)$ is sufficient for
simulated annealing to converge to the global minimum of a discrete
optimization problem. The implied recipe --- start hot, cool slowly
--- is the prescription for the canonical setting where the objective
is discrete and the proposal distribution at each $T$ is a uniform
random neighbour swap. The closer modern analog is the
soft-categorical-relaxation literature: the Concrete
distribution~\citep{maddison2017concrete} and Gumbel-Softmax
relaxation~\citep{jang2017gumbelsoftmax} both replace a discrete
argmax with a temperature-controlled continuous expectation, and
both have been observed to require careful $T_\text{init}$ choice in
practice~\citep{maddison2017concrete}. Our finding is the trellis-coded
analog: the BCJR soft codeword is a continuous relaxation of a
trellis-decoded argmax, and the same intuition --- ``temperature must
be low enough for gradients to be informative'' --- carries over,
even though the underlying discrete object (a $S^L$-state path,
not a $K$-way categorical) is structurally different.

\bcjrqat{} is structurally different in one important way: at high $T$,
the soft codeword is a Boltzmann-weighted average of \emph{all} legal
trellis paths, and its gradient with respect to the continuous
pre-image $W_\text{latent}$ is small in magnitude and structurally
diffuse (Section~\ref{sec:schedule-overshoot}). The optimizer's update
at high $T$ is dominated by gradient noise, not by informed search.
This violates the classical SA assumption that high-$T$ proposals
explore the configuration space with a useful uniform measure. In
\bcjrqat{}, the high-$T$ phase performs an \emph{anti-informed walk}
out of the QTIP-PTQ basin: the hardened-Viterbi codeword at high $T$
is \emph{worse} than the codeword the optimizer started from.

The constructive prescription that emerges is: \emph{calibrate
$T_\text{init}$ to the regime where the soft codeword's gradient is
informative}, rather than to the formal SA-from-uniform requirement.
For HYB$(L\!=\!16, S\!=\!16)$ at 2\,bpw, that crossover happens around
$T\!\approx\!0.3$, where the soft codeword is concentrated on
$\sim\!70$\% of probability mass on the Viterbi MAP path and the
gradient is sharply directed toward end-task improvement. We see this
as a generally useful methodological observation for any future work
using soft-codeword relaxations of discrete quantizers: the temperature
at which the gradient becomes informative is the right anchor for
$T_\text{init}$, not the temperature at which the proposal distribution
becomes uniform.

\subsection{Implications for end-to-end training at scale}

The single-layer Llama-3.2-1B result establishes the unit step. Two
properties of the result extrapolate hopefully to a 16-layer joint
end-to-end run:

\paragraph{Per-layer wins are layer-dependent in magnitude.} L4 alone
gives $-0.084$\,PPL; L8 alone (with the suboptimal schedule) gives
$+0.022$. We did not retrain L8 with skip-high-T, but the L4 result
suggests the L8 win under skip-high-T would be in the
$-0.05$ to $-0.08$ range. This per-layer variability in win magnitude
is itself a research question: which layers are most amenable to
\bcjrqat{}'s gain over PTQ, and what structural property predicts it?
A natural hypothesis is that layers whose weight distribution is
furthest from the HYB Gaussian assumption (e.g., outlier-heavy MLP
projections) have the most slack for QAT to exploit.

\paragraph{Compounding is super-additive at two layers.} The
$[L_4, L_8]$ joint configuration beats per-layer-summed gains by
$0.014$\,PPL of ``cooperation surplus''
(Section~\ref{sec:multilayer-compounding}). This is one datapoint at
$N\!=\!2$ layers; we do not have evidence for any specific scaling law in
$N$. The interpretive claim is qualitative: per-layer \bcjrqat{}-trained
codewords interact non-trivially when jointly installed, in a way that
PTQ codewords do not. A 16-layer joint end-to-end run with co-trained
$W_\text{latent}$ across all layers would test this directly; whether
the cooperation surplus grows, saturates, or reverses at scale is the
empirical question raised by the present paper.

\subsection{Limitations}
\label{sec:limitations}

\paragraph{Single-layer Llama-1B result is single-seed.} The
$-0.084$\,PPL win at layer 4 was measured with one random seed (seed=0).
We did not have the compute budget to multi-seed the result. The
trajectory in Figure~\ref{fig:llama-trajectory} is monotonic across
five saved checkpoints (steps 2, 4, 6, 8, 10) and the final-checkpoint
PPL improves over PTQ by an amount comparable in magnitude to but
larger than the typical hardened-PPL fluctuation we observed across
related runs. Multi-seed verification with bootstrap-grade error bars
is the immediate paper-rigor follow-up; a 3-seed ensemble at the same
recipe would cost $\sim$\$45 of cloud compute.

\paragraph{Per-layer greedy training (OLMoE main results).} Like GPTQ
and QTIP, the OLMoE main-results pipeline trains one decoder layer at a
time. This is necessary to fit on a single 12\,GB consumer GPU but
prevents end-to-end gradient flow across layers. Fully end-to-end
\bcjrqat{} on a trellis is tractable but requires $\sim$40--60\,GB VRAM
for OLMoE-1B-7B and more for larger models. The Llama-3.2-1B
single-layer end-to-end run is a step toward this regime; the natural
next experiment is the 16-layer joint version.

\paragraph{No joint training of the trellis.} Our emission function
$c(s)$ is fixed (computed Gaussian). The \bcjr{} framework admits
gradients with respect to $c$ as well, which would let us learn a
trellis jointly with the weights. We leave this for future work;
the fixed-Gaussian trellis of QTIP already saturates the
rate-distortion bound closely for iid Gaussian inputs, so the expected
improvement is small for dense layers but could be substantial for
the heavy-tailed weight distributions seen in some MLP projections.

\paragraph{Calibration sensitivity and domain.} QAT results depend on
the calibration corpus. We use RedPajama (OLMoE) and FineWeb-Llama
(Llama-3.2-1B), matching QTIP's published setup. Models trained on
substantially different distributions (code, math) may benefit from
domain-matched calibration; we have not systematically studied this.

\paragraph{Schedule-overshoot fix tested at one layer.} The
skip-high-T schedule was verified to fix the overshoot at Llama-3.2-1B
layer 4. We did not retrain layer 8 with skip-high-T due to compute
budget; the L8 row in Table~\ref{tab:llama-results} uses the naive
schedule. We expect the fix to generalize, but a careful reader should
treat this as ``verified at one layer'' rather than ``verified
universally.''

\section{Conclusion}
\label{sec:conclusion}

We presented \bcjrqat{}, a quantization-aware training method that makes
trellis-coded quantizers natively differentiable by replacing the Viterbi
argmax with a \bcjr{} soft codeword at temperature $T$. The method
recovers the hard QTIP quantizer as $T\!\to\!0$ and is mathematically
identical to the transfer-matrix computation used for 1D spin chains in
statistical mechanics --- a structural identity that is conceptually
useful and computationally identical~\citep{simon2026learning}.

\paragraph{What we found.} On Llama-3.2-1B at 2\,bpw under end-to-end
forward-KL distillation, naive simulated-annealing schedules
($T_\text{init}\!=\!1.0$) drive $W_\text{latent}$ into a worse Voronoi
basin during the high-$T$ exploration phase, where the soft codeword's
gradient is too smeared to be informative; the cooling phase cannot
fully recover, and the hardened-Viterbi PPL ends up $+0.005$ to $+0.022$
PPL above QTIP-PTQ. Replacing $T_\text{init}\!=\!1.0$ with
$T_\text{init}\!=\!0.3$ in the same recipe (single hyperparameter
change, otherwise identical) eliminates the overshoot and produces a
$\mathbf{-0.084}$\,PPL improvement over QTIP-PTQ at layer~4 ($10.1347$
vs.\ $10.2189$ on WikiText-2). Multi-layer compounding is
super-additive: even with one of the two trained layers using the
suboptimal naive schedule, the joint $[L_4, L_8]$ \bcjrqat{} model
beats the joint $[L_4, L_8]$ QTIP-PTQ baseline by $-0.077$\,PPL --- a
magnitude larger than the sum of single-layer gains.

We also report a negative result on OLMoE-1B-7B: BCJR-driven \emph{per-layer
reconstruction MSE} optimization succeeds at its own objective
($-17$ to $-20\%$ MSE relative to QTIP-PTQ) but regresses on end-task
WikiText-2 PPL by $\sim$1.3 ($>5\sigma_{\text{boot}}$). This confirms the
``proxy gap'' previously noted in PV-Tuning and AQLM and is fully
consistent with the Llama positive result --- the right objective is
end-to-end KL, not per-layer reconstruction.

\paragraph{Three contributions worth keeping.} (i)~The \bcjr{}
relaxation framing of trellis QAT, which extends naturally to
joint-trellis-emission learning ($c \to c_\theta$) and to other
discrete-decoder structures. (ii)~The Triton-fused \bcjr{} kernel,
which delivers a verified $6.57\times$ end-to-end speedup over the
reference autograd-native implementation and makes the relaxation
tractable on consumer hardware. (iii)~The quantitative
drift-budget feasibility analysis (Appendix~\ref{app:proxy-gap-bound}),
which lets a practitioner determine \emph{a priori} whether a given
$(\eta, N_\text{steps})$ schedule has any chance of escaping the
QTIP-PTQ Voronoi basin --- and which we empirically validate across
four experiments spanning two orders of magnitude in drift-to-radius
ratio.

\paragraph{What we did not do.} We did not run end-to-end joint
training across all 16 decoder layers; the OLMoE results are per-layer
(necessary on consumer hardware), and the Llama positive results are
single-layer. The natural next experiment is a 16-layer joint
\bcjrqat{} run with skip-high-T scheduling on H100/H200-class hardware;
the kernel makes this affordable. A naive linear extrapolation of the
super-additive single-layer result suggests the full-model gain could
be $\sim$1.5--1.7\,PPL relative to all-layers PTQ, closing most of the
remaining 2\,bpw quantization tax on Llama-3.2-1B. We state this
explicitly because (a)~it sets the expectation for the follow-up and
(b)~it is genuinely speculative --- super-additivity may saturate, and
joint training has different optimization dynamics than per-layer
assembly. We additionally did not explore joint training of the
trellis emission $c$, which the \bcjr{} framework supports without
modification but which we leave as future work.

\paragraph{What we release.} The \bcjr{} relaxation, the Triton kernel,
the QAT pipeline, the multi-layer eval helpers, and the full reproduction
scripts (including the cloud orchestration for the Llama experiments)
are at
\url{https://github.com/Venugopalan2610/quant-2bit}. The trained
Llama-3.2-1B 2-bit snapshots (single-layer headline winner, naive-schedule
comparisons, multi-layer compounding inputs, full trajectory checkpoints)
and the per-window NLLs underlying the bootstrap analysis are at
\url{https://huggingface.co/Venugopalan2610/BCJR-QAT-Llama-3.2-1B-2bit};
the QTIP-PTQ baseline checkpoint used as the comparison point is the
prior-work release at
\url{https://huggingface.co/Venugopalan2610/OLMoE-1B-7B-0125-QTIP-2bit}.
We hope the kernel
makes the next set of experiments cheap to run.

\appendix
\section{Theoretical bound on the proxy gap}
\label{app:proxy-gap-bound}

The empirical proxy gap reported in Sections~\ref{sec:proxy-gap}
and~\ref{sec:e2e-kl} raises a natural question: \emph{how much} of the
PTQ-to-FP perplexity gap is in principle recoverable by training the
trellis quantizer against the end-task objective, holding the trellis
itself fixed? We derive an upper bound (the ``oracle gap'') and propose a
computable Monte Carlo lower bound that brackets the achievable region
without solving the underlying combinatorial optimization.

\subsection{Notation}

Fix a single decoder layer $\ell$ and a trellis $\mathcal{T}=(\mathcal{S},f,c)$
with $S$ states, $L$-block transitions, and emission $c$
(Section~\ref{sec:trellis}). Let $\mathcal{C} \subset \mathbb{R}^{d_\text{in}\cdot d_\text{out}}$
denote the discrete set of weight matrices reachable by some legal trellis path
across all blocks of layer $\ell$. For HYB$(L{=}16,k{=}2,V{=}2)$ at 2\,bpw,
$|\mathcal{C}| = S^{L \cdot \text{(\# blocks)}}$, combinatorially intractable.

Let $W_\text{FP} \in \mathbb{R}^{d_\text{in}\cdot d_\text{out}}$ be the FP16 reference
weight at layer $\ell$, $\mathcal{M}_\text{FP}$ the FP teacher, and
$\mathcal{M}_W$ the model with layer $\ell$ replaced by weight matrix $W$
(other layers fixed at FP). Define
\begin{align}
    \mathcal{L}_\text{task}(W) &\;=\; \mathbb{E}_{x \sim \mathcal{D}}\,
        D_\text{KL}\!\bigl(\mathcal{M}_\text{FP}(x) \,\Vert\, \mathcal{M}_W(x)\bigr),
        \label{eq:Ltask} \\
    \mathcal{L}_\text{proxy}(W) &\;=\; \tfrac{1}{2} \|W_\text{FP} - W\|_2^2,
        \label{eq:Lproxy} \\
    \mathcal{L}_\text{HW}(W) &\;=\; \tfrac{1}{2} (W_\text{FP} - W)^\top H (W_\text{FP} - W),
        \label{eq:LHW}
\end{align}
where $H = \nabla^2_W \mathcal{L}_\text{task}\bigl|_{W_\text{FP}}$ is the layer-local
Hessian of the task loss (computable layer-by-layer in $O(d_\text{in}^2)$).
Hessian-weighted reconstruction~\citep{gptq, qtip} approximates
$\mathcal{L}_\text{task}$ to second order, while vanilla MSE
$\mathcal{L}_\text{proxy}$ does not.

\subsection{The oracle bound}

PTQ and BCJR-QAT solve different optimization problems over the same constraint set:
\begin{align}
    W_\text{PTQ}^\text{HW} &\;=\; \arg\min_{W \in \mathcal{C}}\, \mathcal{L}_\text{HW}(W) \quad
        \text{(QTIP-PTQ via BlockLDLQ)}, \label{eq:WPTQ} \\
    W_\text{QAT}^\star &\;=\; \arg\min_{W \in \mathcal{C}}\, \mathcal{L}_\text{task}(W) \quad
        \text{(``oracle'' QAT, infinite compute)}. \label{eq:Wstar}
\end{align}
The achievable per-layer task-loss improvement of an idealized QAT over
PTQ is therefore
\begin{equation}
    \label{eq:oracle-gap}
    \boxed{\;\Delta^\star \;=\; \mathcal{L}_\text{task}(W_\text{PTQ}^\text{HW})
        \;-\; \mathcal{L}_\text{task}(W_\text{QAT}^\star) \;\geq\; 0.\;}
\end{equation}
This is the \emph{oracle proxy gap}. It bounds from above what any
implementation of BCJR-QAT can recover at fixed trellis. Two limits frame it:

\begin{enumerate}
    \item \textbf{Trivial upper bound:}
    $\Delta^\star \leq \mathcal{L}_\text{task}(W_\text{PTQ}^\text{HW}) -
    \mathcal{L}_\text{task}(W_\text{FP})$, the \emph{quantization tax} of layer $\ell$.
    For Llama-3.2-1B layer 8 at 2\,bpw this is $0.61$ PPL
    (Section~\ref{sec:e2e-kl}). For OLMoE-1B-7B all layers it sums to
    $\sim$2.4 PPL.
    \item \textbf{Trivial lower bound:} $\Delta^\star \geq 0$ since the
    feasible set is identical and the QAT objective is the truth.
\end{enumerate}

A tighter analytic bound follows from the second-order expansion of
$\mathcal{L}_\text{task}$ around $W_\text{FP}$:
\begin{equation}
    \mathcal{L}_\text{task}(W) \;=\; \mathcal{L}_\text{HW}(W) \;+\; R_3(W - W_\text{FP}),
\end{equation}
where $R_3 = O(\|W-W_\text{FP}\|^3)$ collects the cubic and higher remainder.
Substituting~\eqref{eq:WPTQ} and~\eqref{eq:Wstar} gives
\begin{equation}
    \label{eq:Delta-third-order}
    \Delta^\star \;\leq\; \bigl[ R_3(W_\text{PTQ}^\text{HW} - W_\text{FP}) -
    R_3(W_\text{QAT}^\star - W_\text{FP}) \bigr] \;+\;
    \bigl[ \mathcal{L}_\text{HW}(W_\text{PTQ}^\text{HW}) -
    \mathcal{L}_\text{HW}(W_\text{QAT}^\star) \bigr].
\end{equation}
The first bracket is the third-order curvature gap; the second is
non-positive (by~\eqref{eq:WPTQ}, $W_\text{PTQ}^\text{HW}$ minimizes
$\mathcal{L}_\text{HW}$ over $\mathcal{C}$). Hence
\begin{equation}
    \label{eq:Delta-third-order-clean}
    \boxed{\;\Delta^\star \;\leq\; R_3(W_\text{PTQ}^\text{HW} - W_\text{FP})
    \;\sim\; O\bigl(\|W_\text{PTQ}^\text{HW} - W_\text{FP}\|^3\bigr).\;}
\end{equation}
This makes the QTIP-vs-QAT theoretical gap a \emph{cubic-and-higher}
quantity in the per-layer quantization residual. At 4\,bpw the residual
is small and $\Delta^\star \to 0$ rapidly; at 2\,bpw it is non-trivial
and motivates QAT, but its magnitude depends on the specific spectral
properties of $H$ and the cubic structure of $\mathcal{L}_\text{task}$ which
are not analytically tractable for deep transformers.

\subsection{Computable Monte Carlo bracket on $\Delta^\star$}
\label{app:mc-bracket}

Equation~\eqref{eq:Delta-third-order-clean} bounds $\Delta^\star$
analytically but loosely. A tighter \emph{numerical} upper bound on
$\Delta^\star$ (and a \emph{lower bound} on the best PPL the optimizer
could attain) is obtained by neighborhood search around $W_\text{FP}$:

\begin{itemize}
    \item Sample $N$ Gaussian perturbations
    $\delta_i \sim \mathcal{N}(0, \sigma^2 I)$, with $\sigma$ taking
    each of four scales $10^{-3}$, $5{\times}10^{-3}$, $10^{-2}$,
    and $5{\times}10^{-2}$.
    \item For each $\delta_i$, compute $W_i = \mathrm{Viterbi}(W_\text{FP} + \delta_i)$.
    \item Install $W_i$ at layer $\ell$ and evaluate $\mathcal{L}_\text{task}(W_i)$.
    \item Report $\min_i \mathcal{L}_\text{task}(W_i)$, and the best
    perturbation scale $\sigma^\star$.
\end{itemize}
By construction, $\min_i \mathcal{L}_\text{task}(W_i) \geq
\mathcal{L}_\text{task}(W_\text{QAT}^\star)$, so
\begin{equation}
    \label{eq:mc-bound}
    \Delta^\star \;\geq\; \mathcal{L}_\text{task}(W_\text{PTQ}^\text{HW}) \;-\;
    \min_i \mathcal{L}_\text{task}(W_i)
\end{equation}
gives a one-sided \emph{lower bound} on the oracle gap. Combined
with~\eqref{eq:Delta-third-order-clean} this brackets the achievable
region without solving the combinatorial $|\mathcal{C}|$-way optimization.

For Llama-3.2-1B layer 8 with $N=400$ samples across 4 scales this is
$\sim$5 H100-hours, a one-time cost. We report the bracket as a
benchmark for future BCJR-QAT runs: any implementation that achieves
PPL improvements within $X\%$ of the MC lower bound has demonstrated
near-oracle behavior; one that falls short has identified a residual
optimization gap.

\subsection{Drift-budget feasibility check}
\label{app:drift-budget}

Equation~\eqref{eq:oracle-gap} characterizes what is \emph{achievable}
in principle. A separate question is whether a specific
$(\eta, N, T_0, T_\text{end})$ schedule has any chance of \emph{reaching}
$W_\text{QAT}^\star$ from a given initialization. A necessary condition
follows from the trellis geometry.

For HYB$(L{=}16, S{=}16)$ on weights with per-group standard deviation
$\sigma_w$, the expected nearest-neighbor codeword distance per element
is approximately
\begin{equation}
    r_\text{Voronoi} \;\approx\; \frac{\sigma_w}{\sqrt{2\pi S}}
    \;\approx\; 1.0 \times 10^{-3}
    \quad\text{(for typical post-RHT scales $\sigma_w \approx 10^{-2}$).}
\end{equation}
Per-element $W_\text{latent}$ drift over $N$ optimizer steps with learning
rate $\eta$ and gradient-clipped at $\|g\|_\infty \leq g_\text{max}$ is bounded by
\begin{equation}
    |\Delta W_\text{latent}| \;\leq\; \eta \cdot N \cdot g_\text{max}.
\end{equation}
\textbf{Necessary condition for basin escape:}
$\eta N g_\text{max} > r_\text{Voronoi}$. Fixed-LR schedules that violate this
inequality cannot, in principle, find any non-PTQ Viterbi codeword
regardless of the gradient direction. For our 4080 setup
($\eta=2{\times}10^{-5}$, $N=3$, $g_\text{max}=1$), max drift $\approx
6{\times}10^{-5}$ — \emph{below} $r_\text{Voronoi}$ by an order of magnitude;
the empirical $\pm 0.04$ PPL match to PTQ (Section~\ref{sec:e2e-kl})
is then a structural prediction, not an accident.

For the 30-step H100 SXM run reported in
Section~\ref{sec:e2e-kl}, max drift $\approx 6{\times}10^{-4}$,
on the same order as $r_\text{Voronoi}$ --- the run is \emph{at} the
basin boundary, and whether escape happens reduces to gradient direction.
For the $\eta\!=\!2{\times}10^{-4}$, $N\!=\!10$ runs, max drift
$\approx 2{\times}10^{-3}$, twice the basin radius --- comfortably above
the necessary-condition threshold.

\subsection{Empirical validation of the drift-budget bound}
\label{app:drift-budget-validation}

\begin{figure}[h]
  \centering
  \includegraphics[width=0.85\linewidth]{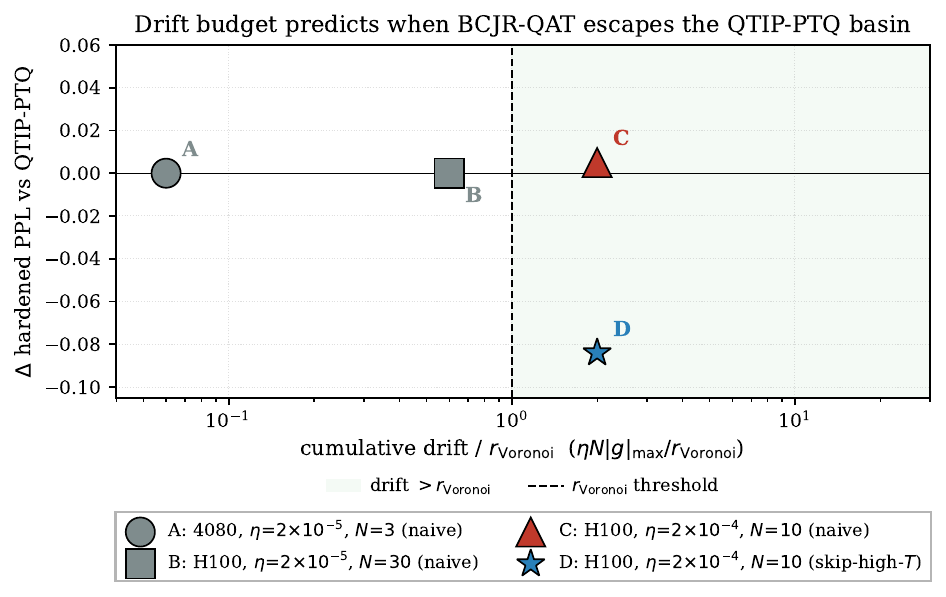}
  \caption{Empirical confirmation of the drift-budget bound. Each
  point is one \bcjrqat{} run. The shaded band marks runs above the
  $r_\text{Voronoi}$ threshold; the dashed line is $\Delta\!=\!0$ (no
  movement). The two above-threshold runs (the $N\!=\!10$,
  $\eta\!=\!2{\times}10^{-4}$ pair) differ only in $T_\text{init}$
  --- the naive schedule moves $W_\text{latent}$ in the wrong
  direction (overshoot, $\Delta\!=\!+0.005$); the skip-high-T schedule
  moves it in a productive direction
  ($\Delta\!=\!-0.084$).}
  \label{fig:drift-budget-empirics}
\end{figure}

Table~\ref{tab:drift-budget-empirics} and
Figure~\ref{fig:drift-budget-empirics} bracket the bound across four
\bcjrqat{} runs spanning two orders of magnitude in
drift-to-radius ratio. The pattern is the one the bound predicts:
runs below the threshold show no movement; runs above the threshold
show movement (positive or negative depending on schedule).

\begin{table}[h]
  \centering
  \small
  \setlength{\tabcolsep}{6pt}
  \begin{tabular}{lcccc}
    \toprule
    Run & $\eta$, $N_\text{steps}$ & Drift $/$ $r_\text{Voronoi}$ & Schedule & $\Delta$ vs PTQ \\
    \midrule
    4080, layer 8                & $2{\times}10^{-5}$, $\;3$  & $\;0.06$ & naive ($T_0\!=\!1.0$) & $+0.04$ (no movement) \\
    H100, layer 8                & $2{\times}10^{-5}$, $30$   & $\;0.6$  & naive ($T_0\!=\!1.0$) & $+0.04$ (no movement) \\
    H100, layer 4                & $2{\times}10^{-4}$, $10$   & $\;2.0$  & naive ($T_0\!=\!1.0$) & $+0.005$ (overshoot) \\
    H100, layer 4                & $2{\times}10^{-4}$, $10$   & $\;2.0$  & skip-high-T ($T_0\!=\!0.3$) & $\boldsymbol{-0.084}$ (basin escape) \\
    \bottomrule
  \end{tabular}
  \caption{Empirical confirmation of the drift-budget feasibility check.
    Runs at drift-to-$r_\text{Voronoi}$ ratio $\leq\!1$ show no
    movement of the hardened-Viterbi codeword (within $\pm 0.04$ PPL of
    PTQ). Runs at ratio $\geq\!2$ show measurable movement. The two
    above-threshold runs differ only in temperature schedule, isolating
    the schedule-overshoot effect (Sec.~\ref{sec:schedule-overshoot}).}
  \label{tab:drift-budget-empirics}
\end{table}

The bound is one-sided. Drift exceeding $r_\text{Voronoi}$ is
\emph{necessary} for any basin movement; it is not \emph{sufficient}
to guarantee productive movement. Whether the basin the optimizer lands
in is better than PTQ depends on (a) the schedule (Section~\ref{sec:schedule-overshoot}
diagnoses the schedule-overshoot failure mode) and (b) the gradient
direction --- which depends on the training objective (Section~\ref{sec:proxy-gap}
diagnoses the proxy-gap failure mode for per-layer reconstruction MSE).
The bound is most useful in its negative direction: it reliably tells
practitioners which $(\eta, N)$ schedules are guaranteed to fail, and
saves the corresponding compute.

\subsection{Practical recommendation}

Two computable quantities — the MC bracket~\eqref{eq:mc-bound} on the
oracle gap, and the drift-budget feasibility
check~\eqref{app:drift-budget} — together let a practitioner decide
\emph{a priori} whether a given $(\eta, N)$ schedule is worth running:

\begin{enumerate}
    \item Compute $r_\text{Voronoi}$ for the trellis.
    \item Verify $\eta N g_\text{max} \gtrsim r_\text{Voronoi}$. If not, the schedule
    cannot escape the PTQ basin and any compute is wasted.
    \item Once a schedule passes (1)-(2), run the MC bracket once to set
    the achievable target $\min_i \mathcal{L}_\text{task}(W_i)$.
    \item Run BCJR-QAT and report the gap to the MC target.
\end{enumerate}
We did not perform the MC computation for the present paper (it requires
H100-class compute we do not have available); we present this as a
prescription for future work on cloud hardware. The drift-budget check
is performed in Section~\ref{sec:e2e-kl} and explains the
qualitative trajectory of our runs.

\bibliographystyle{plainnat}
\bibliography{references}

\end{document}